\documentclass{ifacconf}

\usepackage{graphicx}      
\usepackage{natbib}        
\usepackage{subcaption}
\usepackage{url}
\begin{document}
\begin{frontmatter}

\title{Deep Learning for Model-Free Prediction of  Thermal States of Robot Joint Motors} 


\author[First]{Trung Kien La} 
\author[First]{Eric Guiffo Kaigom} 

\address[First]{Department of Computer Science \& Engineering,
Frankfurt University of Applied Sciences, 60318 Frankfurt am Main, Germany (e-mails: trung.la@stud.fra-uas.de; kaigom@fb2.fra-uas.de).}

\begin{abstract}                
	In this work, deep neural networks made up of multiple hidden Long Short-Term Memory (LSTM) and Feedforward  layers are trained to predict the  thermal behavior of the joint motors of robot manipulators. A model-free and scalable approach  is adopted. It accommodates   complexity and uncertainty challenges stemming from the derivation, identification, and validation of a large number of  parameters of an approximation model that is hardly available. To this end, sensed joint torques are  collected  and processed  to foresee the thermal behavior of joint motors. Promising prediction results of the machine learning based capture of the temperature dynamics of joint motors of a redundant robot with seven joints are presented. 
\end{abstract}


\end{frontmatter}

\section{Introduction}
\begin{figure}[t]
	\centerline{\includegraphics[width=0.9\columnwidth]{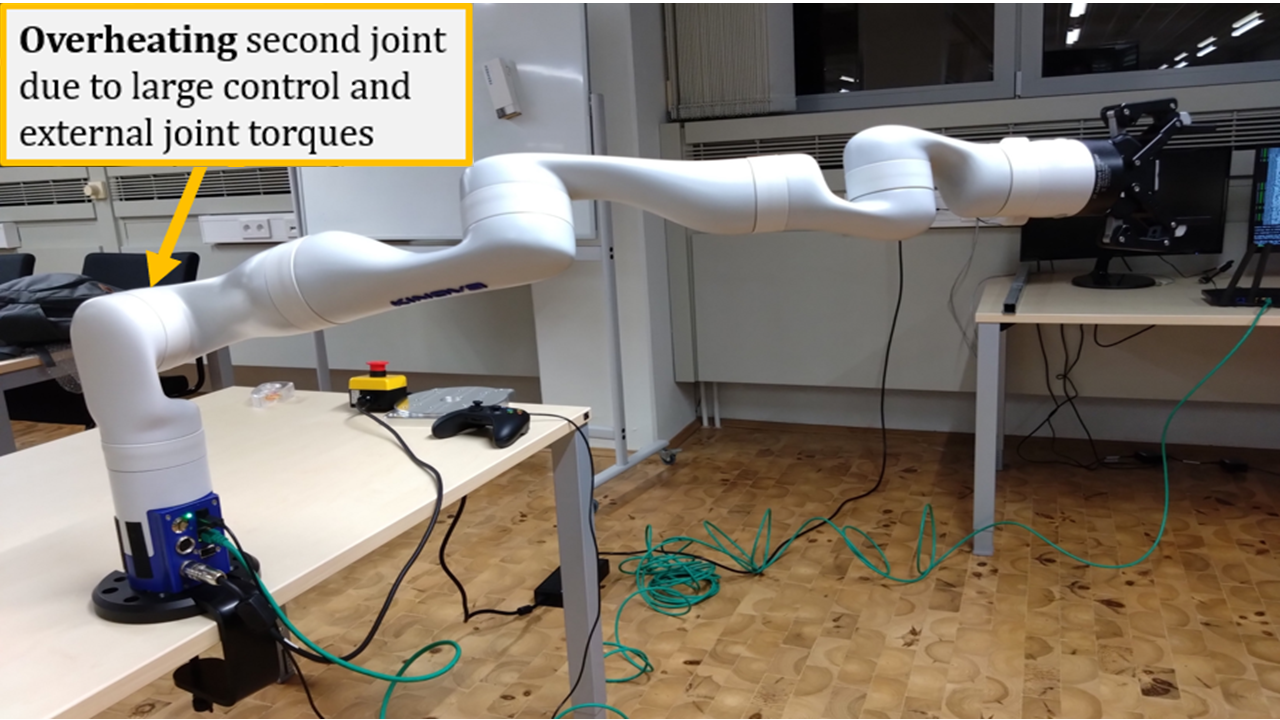}}
	\caption{Posture increasing the temperature of the 2. motor.}
	\label{fig:intro}
\end{figure}

Robots are commonly used to  achieve repetitive and hazardous tasks in industry and society. Meanwhile, they increasingly and skillfully assist and augment humans. Included are  applications with a pronounced level of physical interactions between humans and robots, such as using a robot as a companion (\cite{basha2025robotic}), home-helper and caregiver (\cite{tsui2025exploring},\cite{gkiolnta2025challenges}), as well as a prosthesis (\cite{kim2025mode}). In this respect, high payload manipulations, large joint accelerations, and motions with specific configurations (see, e.g., Fig. \ref{fig:intro}) can induce the overheating of joint motors of robots. Excessive motor temperatures are detrimental in many ways. They can accelerate the degradation of insulation materials and reduce motor efficiency (\cite{yehorov2025study}) along with jeopardizing the positioning accuracy of the robot due to axial  deformations and drifts (\cite{soga2024skillful}).

Most robot manufacturers, including Franka, Kinova,  and KUKA, offer built-in functions to shut down the robot once a critical temperature threshold is attained. Whereas this functionality is advantageous to preserve the performance and reliability of motors and surrounding electronic components, an undesired shutdown tends to compromise the robot availability for production and assistance purposes. This situation gets exacerbated as the robot is not equipped with mechanical brakes, as in Fig. \ref{fig:intro}. In this case, critical collisions with the environment might occur, endangering human beings or leading to hardware (i.e., robot, workpiece, workcell, etc) damages. Furthermore, thermal burns represent  not only  a severe safety issue in physical human-robot-interaction, but also a hindrance for elevated user experience that is necessary to engage and \mbox{sustain a symbiosis between humans and robots.}
\begin{figure*}[t!]
	\begin{center}
		\includegraphics[width=0.76\textwidth]{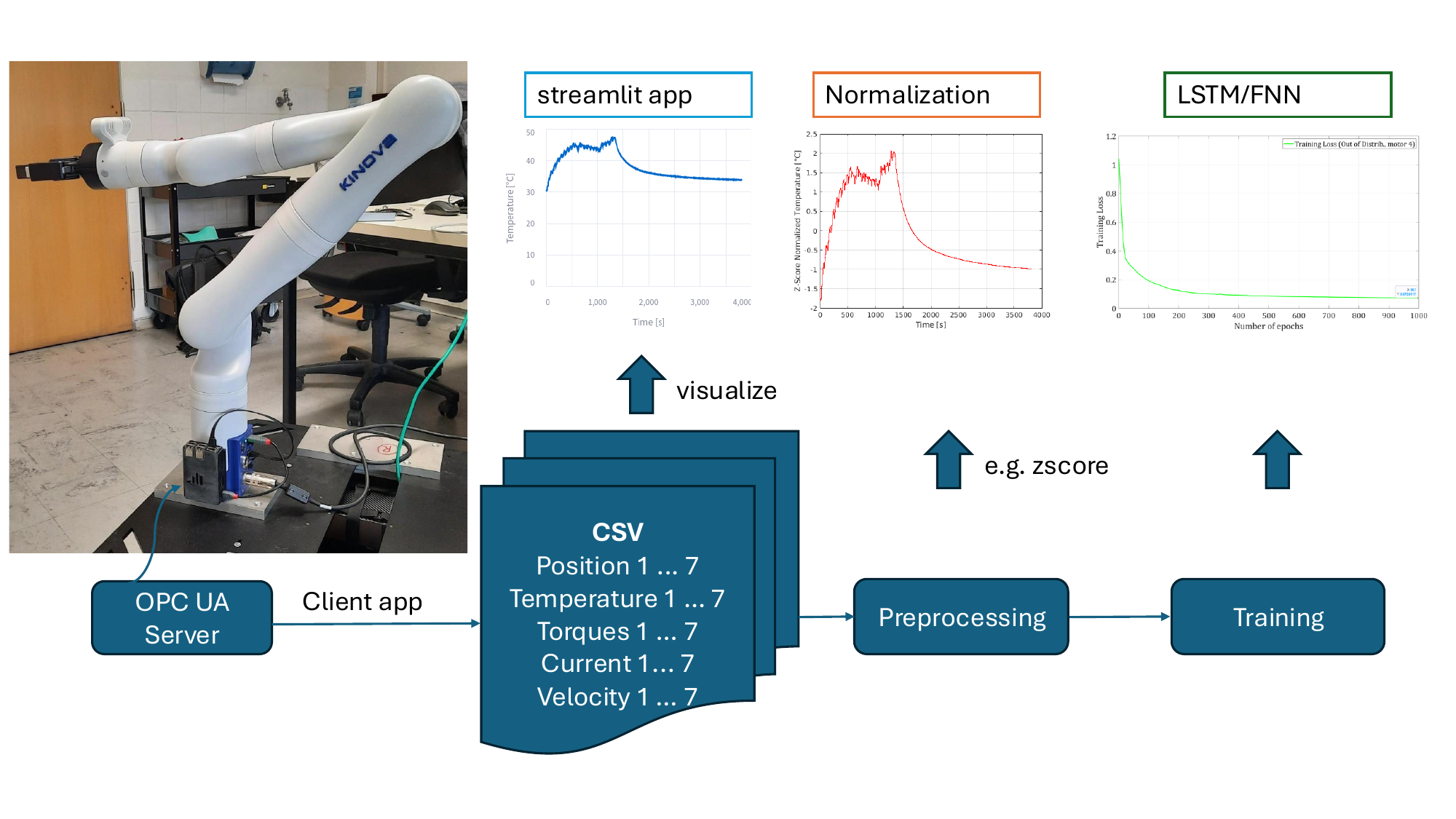} 
		\caption{Data collection from the Kinova robot (l.h.s) via an OPC UA Server and simplified processing pipeline (r.h.s).} 
		\label{fig:DataAquisition}
	\end{center}
\end{figure*}
Predicting the thermal behavior of robot joints is therefore an essential step toward the development of countermeasures that help anticipate overheating, preserve the robot availability,  prolong its lifetime, and improve its usability. Industry 4.0, industry 5.0, together with society 6.0 are likely to benefit from this capability. In fact, the thermal prediction propels operational efficiency through its significant contribution to the design of thermally uncritical trajectories that  preserve the motor performance and thus robot availability in fully automated high-speed low-time-to-market manufacturing. This skill is missing in current industrial applications. Endowing robots with a thermal management based upon machine learning is useful to accommodate uncertainties emanating from \textit{unseen} events. These include the highly dynamic robotized human-centered assistance for smart living and social well-being. These goals require an approach that turns robot diversity and application uncertainties into competitive advantages in terms \mbox{of flexibility,  transferability, and scalability.}

This work predicts  the thermal behavior of joint motors. The prediction approach is
\begin{itemize}
	\item data-driven, paving the ground for an  insightful, non-invasive, and inclusive operationalization in real-time. 
	\item model-free through the development of a deep learning-based framework that leverages multiple hybrid layers to capture the unknown dynamics between joint actuation and motor temperature. No system-related parametric or actuation profile assumptions are made. Model complexity is thereby avoided.  Generalization to new events and transferability to other robot types {regardless of the number and type of joint motors are fostered.}
	\item evaluated on data collected from a redundant Kinova Gen 3 robot with seven joints (see Fig.~\ref{fig:DataAquisition}). Practical advantages of data normalization are highlighted.
\end{itemize}

\section{Related Work}
\label{chap:relatedworks}
Predicting the temperature behavior of robot motors has attracted attention in the recent years. %
\cite{afaq2023intelligent} focus on thermal management of robotic applications under extreme temperatures. Electronics heating and cooling are considered. A temperature control driven by fuzzy logic is developed and demonstrated to this end. Decreases of extreme high temperature from $50^{\circ}$ to $8^{\circ}$ are shown. Fan-based forced convection is used to cool electronics. Excessive internal temperatures in a permanent magnet synchronous motor (PMSM)  taking non-stationary loads, which might lead to a reduction of its life time, is addressed in \cite{chen2024lifetime}. An accelerated degradation model is derived to evaluate the reliability function and predict the lifetime of the PMSM under thermal stress.  Geometric backlash and temperature-related drift errors in joints of industrial robots are compensated in \cite{sigron2023compensation}. A model that reflects the thermal expansion of links is developed and used for thermal expansion correction.  LSTM Neural
Networks (\cite{he2024rotor})  and   Pseudo-Siamese Nested LSTM (\cite{cai2021temperature}) are employed to predict the temperature in Permanent Magnet Synchronous Motors. A trapezoidal torque profile is employed in \cite{he2024rotor} whereas the torque dynamics is not released in (\cite{cai2021temperature}). A thermal recovery of robot joints is achieved in  \cite{jorgensen2019thermal}. To this end, the thermal dynamics is captured as a first order ordinary differential equation subject to constant positive and negative step-like profiles of joint torques. The exponential-based dynamics of the temperature behavior is derived.  A parameter identification is carried out to demonstrate the performance of the model for step-like joint torques. 

Contributions mentioned thus far are mostly model-driven. They fit  with specific robots provided that parameters have been  identified in advance. However, parameter identification requires   noise robustness and low  sensitivity (\cite{zhang2024model}, \cite{de2024non}), which is a time consuming analysis task  prone to additional uncertainties due to unseen$\slash$unmodeled$\slash$truncated dynamics (\cite{shang2024general}). Sometimes, such a process must  be repeated from scratch for a given new robot, which inhibits quick and large scale automation involving multiple robots in terms of complexity, workload, and costs. As the robot is hardly accessible, such as in space servicing,  identification tasks might be hard to complete because of limited access to pertinent (e.g., excitation) data. Autonomous task completion without human interventions, as expected by Industry 6.0, calls for machine learning-embedded solutions (\cite{carayannis2024toward}) that can extend the robot skills for self-condition monitoring. The approach proposed in this work also falls into this category. Robot datasets available from standard APIs are harnessed to predict the thermal behavior of its joint motors. Trained machine learning models can be run by  services of  digital twins (see \cite{kaigom2023metarobotics} and \cite{kaigom2020value}) with which the physical robot is embedded to detect and anticipate  detrimental thermal issues. In contrast to related works, no restriction is made  on  robot types, {number of joints, and actuation profiles.}

\section{Method}

\subsection{System overview}
The full state of our experimental Kinova Gen 3 ultra-lightweight robot with seven degrees-of-freedom (\cite{kinova}) is recorded using a specially developed OPC Unified Architecture server (\cite{Girke}). 
Positions, temperatures, torques, velocities and currents of each of the seven joints are stored, via a client application, as parameters in CSV files and subsequently processed for the neural network trainings, see Fig.~\ref{fig:DataAquisition}.
To cover a wide range of robotic movements, both randomly generated joint angle trajectories and predefined Cartesian trajectories for pick and place tasks, for example, are generated using the Kortex API (\cite{kortexAPI}). 
The trajectories are initially tracked at varying speeds and with different payloads. The recording duration of each set of trajectories is also varied in order to analyze the temperature rise and the cooling behaviour of the joints. 
For cooling, the robot was positioned in a vertical position, as this joint configuration imposes minimal load torques on the joints. Movements without intentional cooling are also conducted sequentially to include as realistic and diversified conditions as possible.
 Randomized joint position are constrained within minimum and maximum limits to avoid potential collisions with the environment. 
Completed experiments indicate that  the second and fourth joint exhibit the highest temperature increases. 

\subsection{Temperature profile approximation}
\begin{figure}[t]
	\begin{center}
		\includegraphics[width=7.5cm]{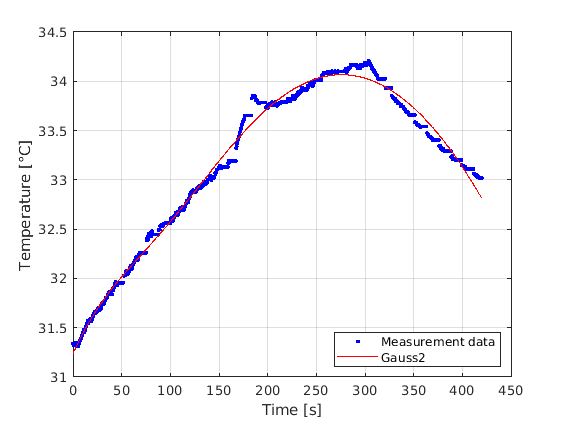}    
		\caption{Approximation of the temperature profile of the 4th joint with the Gauss2 function.} 
		\label{fig:Gauss2_legend}
	\end{center}
\end{figure}

The temperature profile can be represented approximately with the Gaussian model (i.e. Gauss2) as shown in Fig.~\ref{fig:Gauss2_legend} and outlined in eq.~(\ref{eq:gauss2}):
\begin{equation} \label{eq:gauss2}
  \begin{array}{l}
  f(x) = a_1 \cdot \exp\left( -\left( \frac{x - b_1}{c_1} \right)^2 \right) \\
  \quad + a_2 \cdot \exp\left( -\left( \frac{x - b_2}{c_2} \right)^2 \right)
  \end{array}
  \end{equation}
Coefficients and metrics for the  profile in Fig.~\ref{fig:Gauss2_legend} are:
\[
a_1 = 34.07, \quad b_1 = 276, \quad  c_1 = 743.2,
\]

\[
a_2 = 1.668, \quad b_2 = -26.71, \quad c_2 = 103
\]

Root Mean Squared Error (RMSE): 0.081294\\
$R^2$ (coefficient of determination): 0.9897\\

Although the Gauss2 approximation is accurate, it has some limitations. 
The coefficients vary for each temperature profile and must be recalculated. 
Neural networks offer a more generalizable and robust approximation.

\subsection{Feedforward and LSTM neural networks}
Feedforward neural networks (FNN) shift data from the input to the output of a neural network. Backpropagation is run to optimize the neural weights. FNN have a simplistic structure and are limited in their ability to incorporate past values. 
Recurrent neural networks (RNN) therefore have feedback loops to store past values via hidden states. This allows them to understand long-term dependencies, which results in more stable predictions.
However, RNNs lose old information over time. With LSTMs, long-term information is stored more effectively. They feature cell state for long-term memory and hidden state for short-term memory. In addition, the storage and removal of information is controlled via four gates (Input Gate, Forget Gate, Cell Gate, Output Gate).  
They are therefore used for predictions where a large number of past states have an influence on future states (\cite{Ljung}).

Table~\ref{tb:FFLSTM} compares  neural networks integrated in this work to benefit from their respective strengths. In essence,
\begin{table}[b!]
  \begin{center}
  \caption{Comparison between Feedforward and LSTM neural networks}\label{tb:FFLSTM}
  \resizebox{0.8\columnwidth}{!}{  
  \begin{tabular}{lcc}
  \hline
  {Feature} & {FNN} & {LSTM} \\
  \hline
  Handles Sequential Data & No & Yes \\
  Computational Complexity & Low & High \\
  Training Time & Fast & Slow \\
  Memory Requirements & Low & High \\
  Suitable for Static Inputs & Yes & No \\
  Suitable for Time-Series Data & No & Yes \\
  \hline
  \end{tabular}
  }
  \end{center}
\end{table}

\begin{figure*}[t!]
	\centering
	\begin{subfigure}[b]{1.\columnwidth}
		\centering
		\includegraphics[height=2.1in]{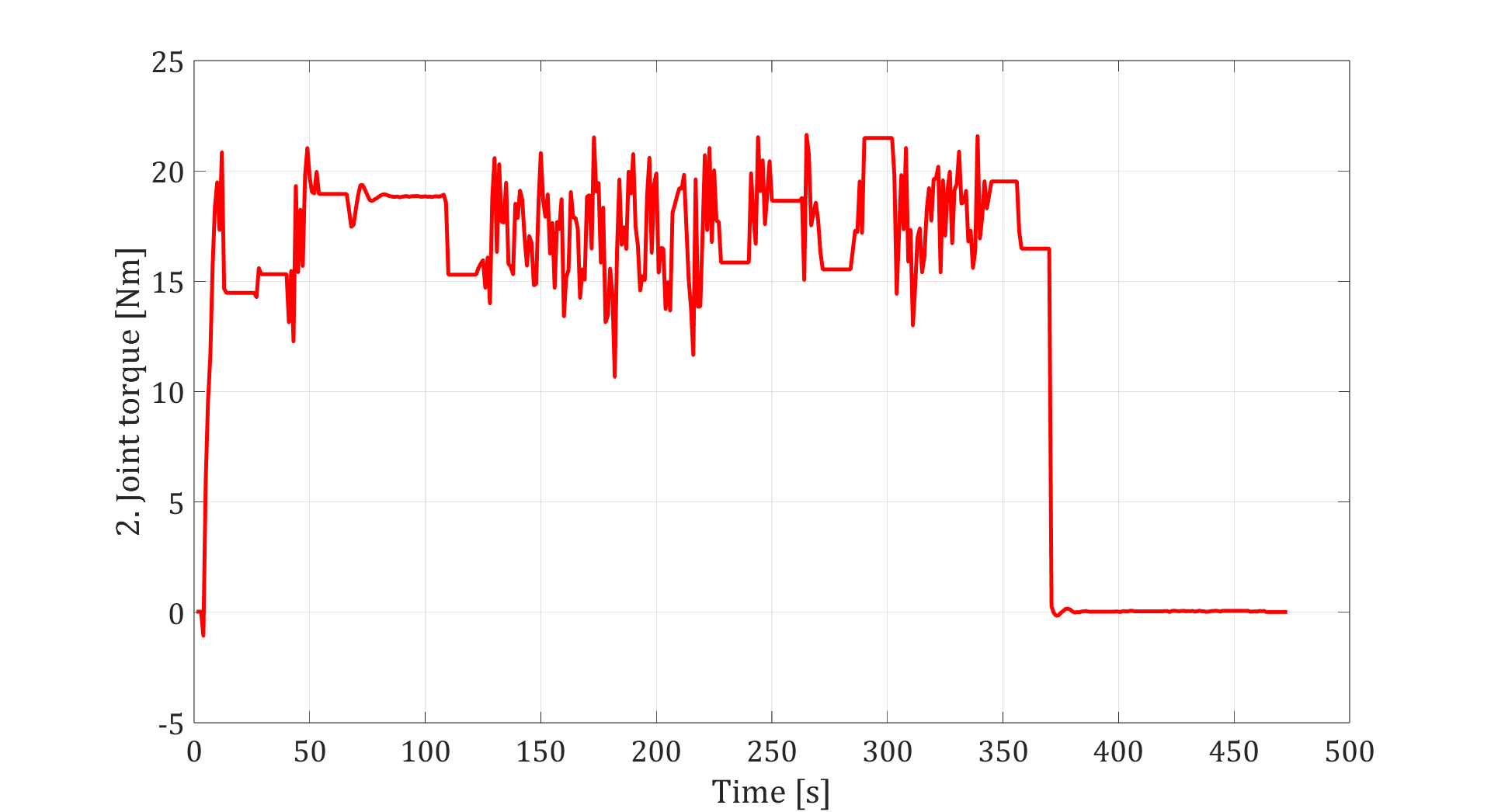}
		\caption{Torque profile of the 2. motor.}
	\end{subfigure}%
	~ 
	\begin{subfigure}[b]{1.\columnwidth}
		\centering
		\includegraphics[height=2.1in]{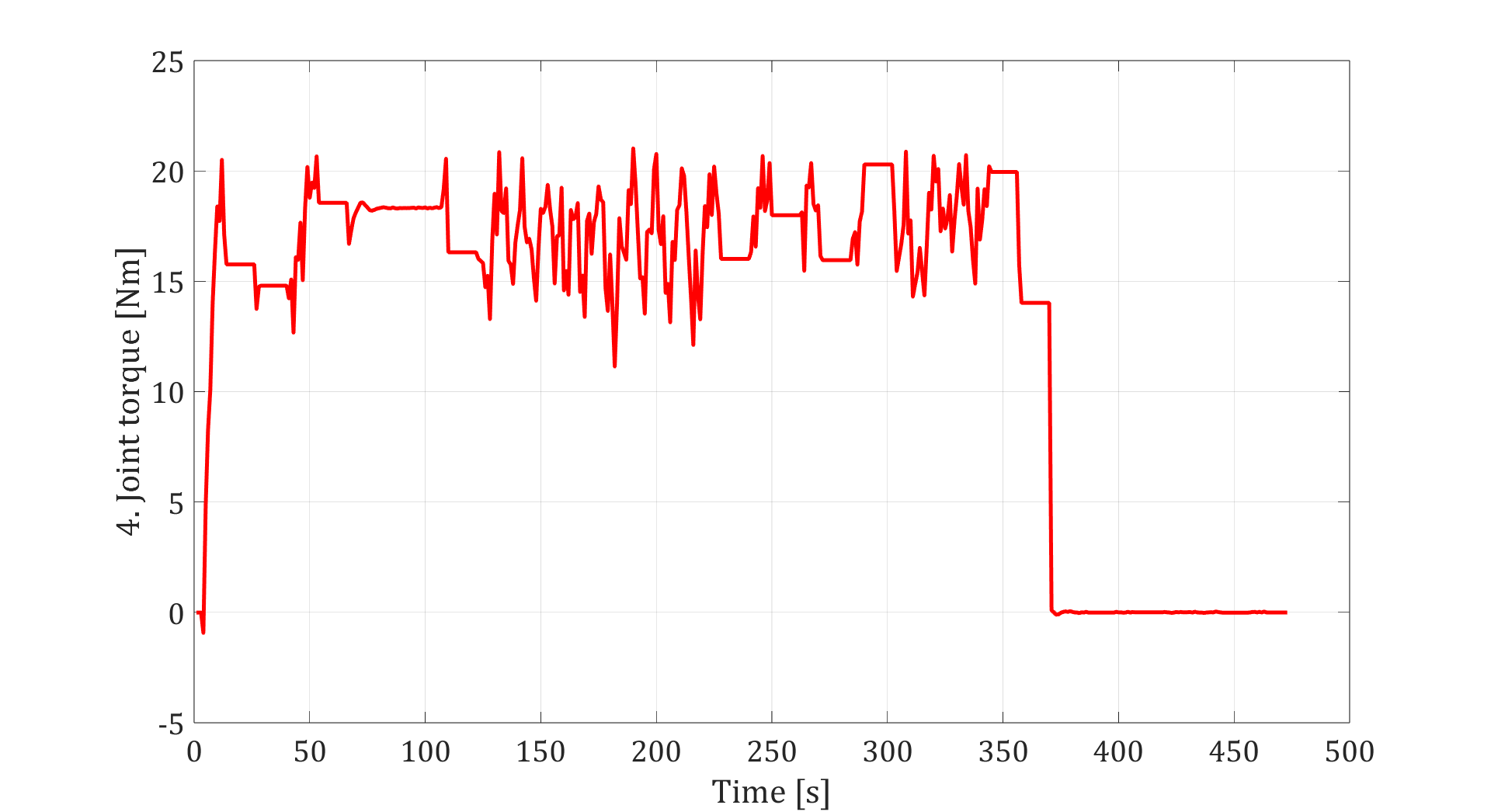}
		\caption{Torque profile of the 4. motor.}
	\end{subfigure}
	\caption{Two different non-trivial  torque profiles of the robot in Fig.~\ref{fig:intro}. Observe that the profiles go beyond step functions. }
	\label{jointtorque}
\end{figure*}

\begin{figure*}[t!]
	\centering
	\begin{subfigure}[b]{1.\columnwidth}
		\centering
		\includegraphics[height=2.1in]{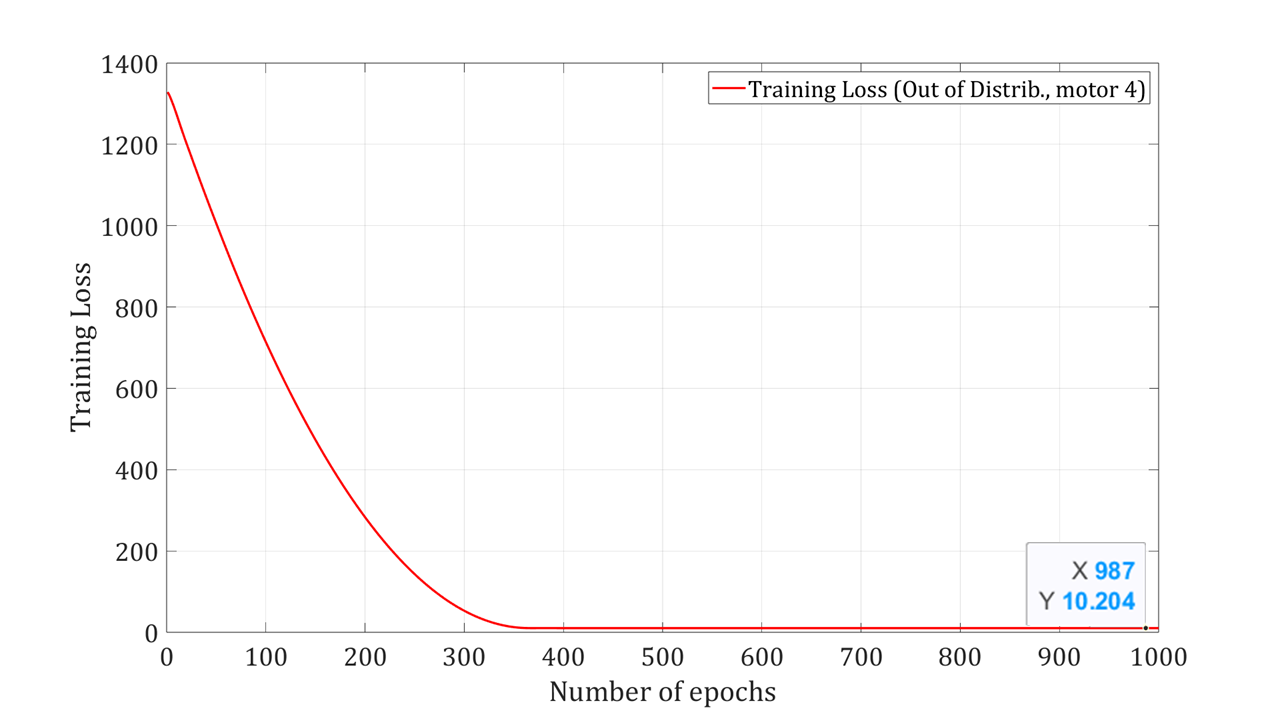}
		\caption{Training loss without data normalization.}
	\end{subfigure}%
	~ 
	\begin{subfigure}[b]{1.\columnwidth}
		\centering
		\includegraphics[height=2.1in]{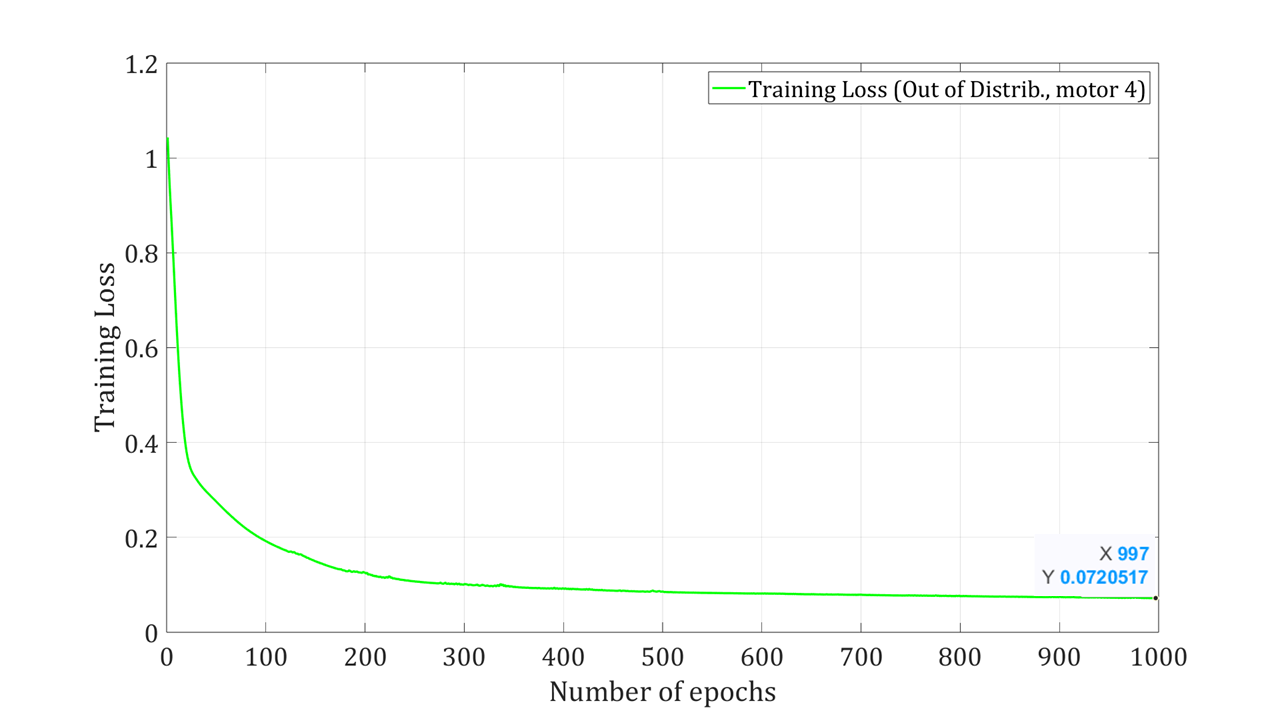}
		\caption{Training loss with z-score data normalization.}
	\end{subfigure}
	\caption{Enhanced  convergence velocity and effectiveness of the training loss through data normalization.}
	\label{trainingloss}
\end{figure*}

\begin{itemize}
  \item FNNs disregard historical information and consider each input independently.  LSTM networks are able to store the context of previous points in time (sequential data) through memory cells and use this information for future predictions (\cite{Liu}).
  \item FNNs can be trained with lower computational power and in a shorter time compared to LSTMs, which consists of more complex structures (e.g., states and gates). LSTMs store and update past information.
  \item Joint temperatures change as a result of continuous loads and the temporal dynamics of joint torques. The time-dependency can be captured by LSTMs. They are specifically designed for time series. FNNs are more suitable for regression problems on static data points where a disturbance or modifying of the temporal dynamics is non-critical.
\end{itemize}
We implement and train a deep neural network that combines LSTM as a feature-extractor for time series dependencies in the input torque data and multiple Feedforward layers to take advantage of the extracted features for the subsequent temperature prediction at the network output.

\begin{figure*}[t!]
	\begin{subfigure}[t]{0.5\textwidth} 
		\centering 
		\includegraphics[height=2.1in, width=\linewidth, keepaspectratio]{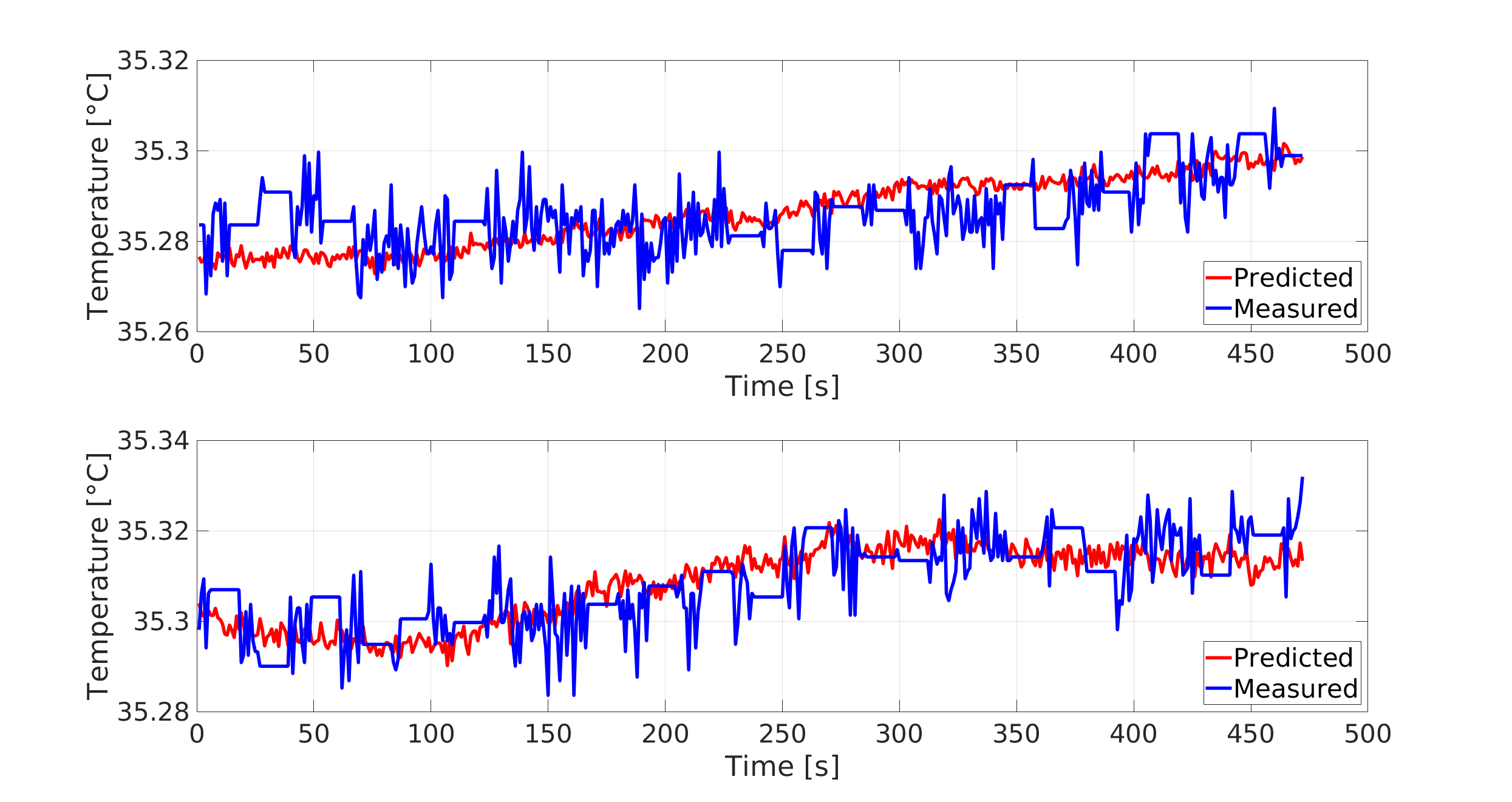}
		\caption{Motor 1: Different predictions with \textbf{unseen} joint torques.}
	\end{subfigure}%
	\begin{subfigure}[t]{0.5\textwidth}
		\centering
		\includegraphics[height=2.1in, width=\linewidth, keepaspectratio]{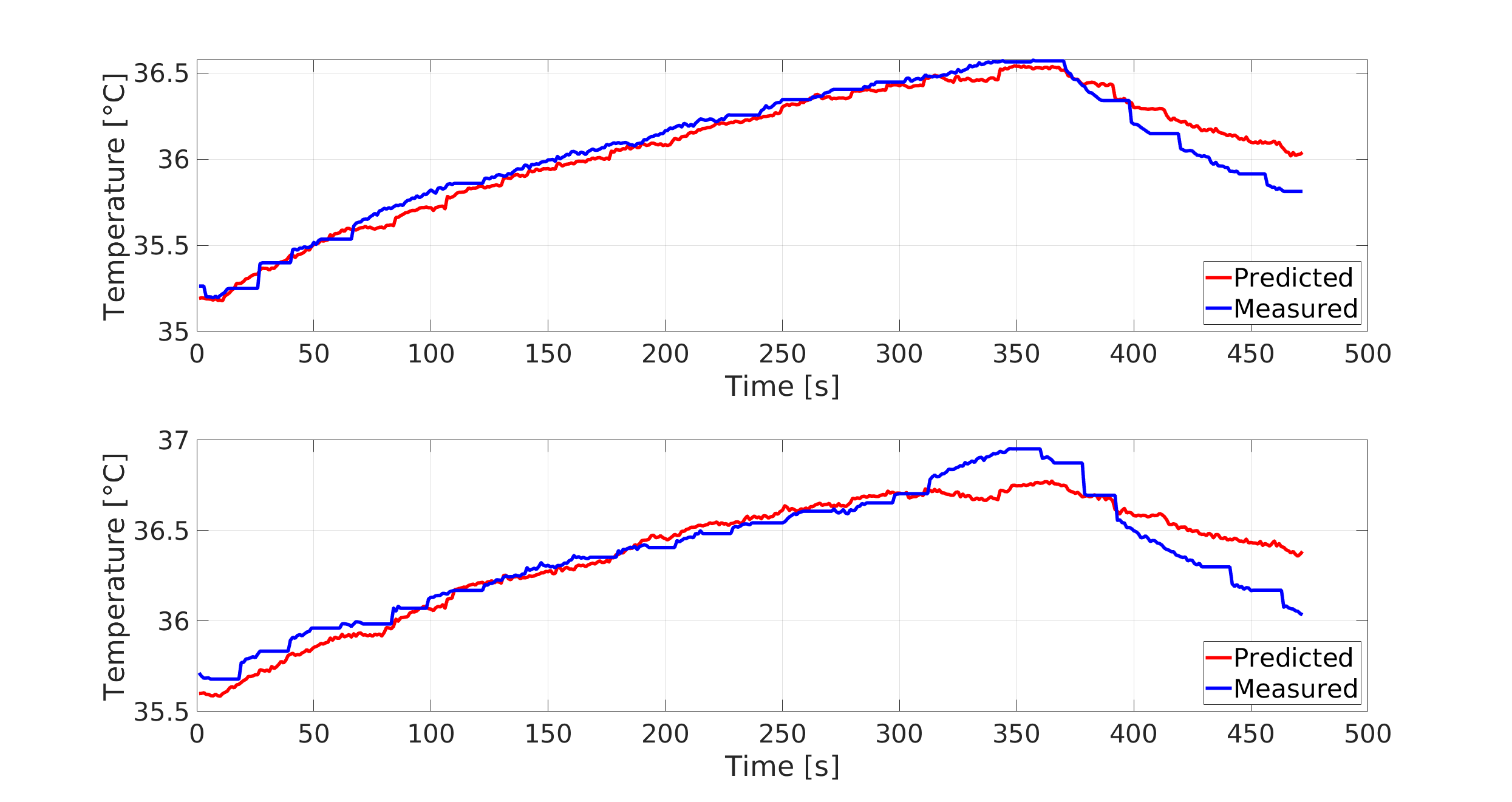}
		\caption{Motor 2: Different predictions with \textbf{unseen} joint torques.}
	\end{subfigure}

    \vspace{1ex} 

    \begin{subfigure}[t]{0.5\textwidth}
		\centering
		\includegraphics[height=2.1in, width=\linewidth, keepaspectratio]{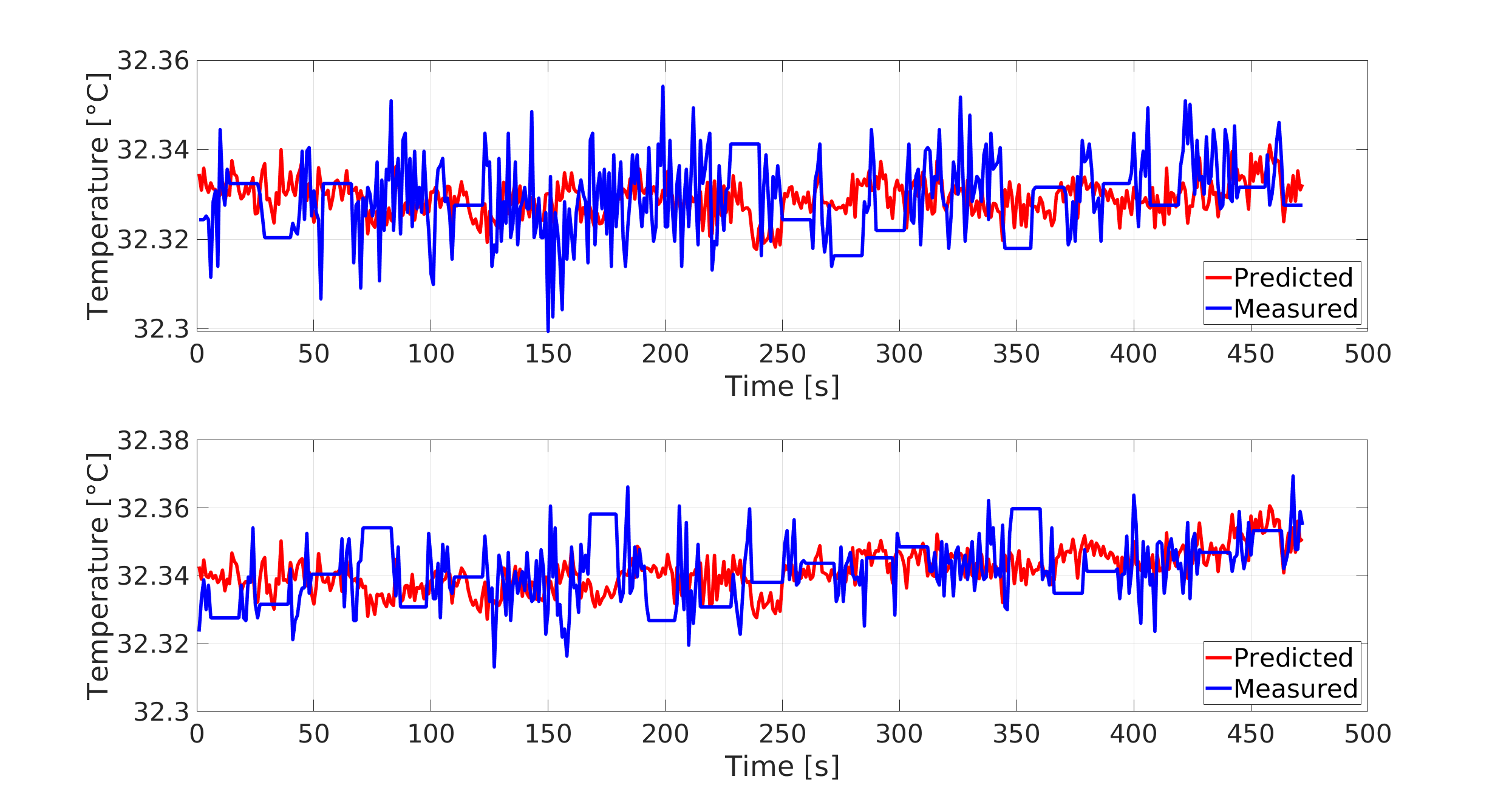}
		\caption{Motor 3: Different predictions with \textbf{unseen} joint torques.}
	\end{subfigure}%
	\begin{subfigure}[t]{0.5\textwidth}
		\centering
		\includegraphics[height=2.1in, width=\linewidth, keepaspectratio]{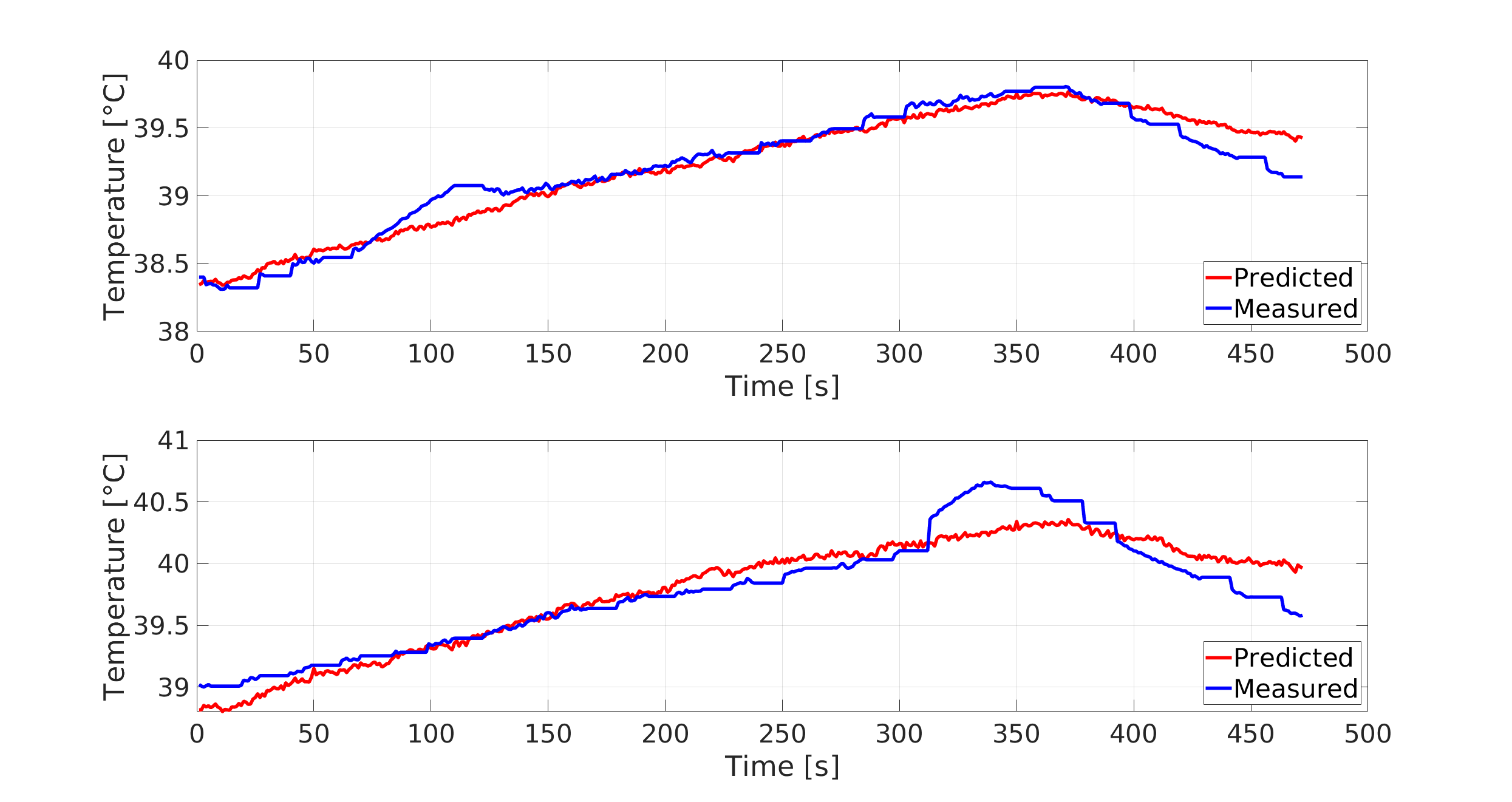}
		\caption{Motor 4: Different predictions with \textbf{unseen} joint torques.}
	\end{subfigure}

    \vspace{1ex} 

	\begin{subfigure}[t]{0.5\textwidth}
		\centering
		\includegraphics[height=2.1in, width=\linewidth, keepaspectratio]{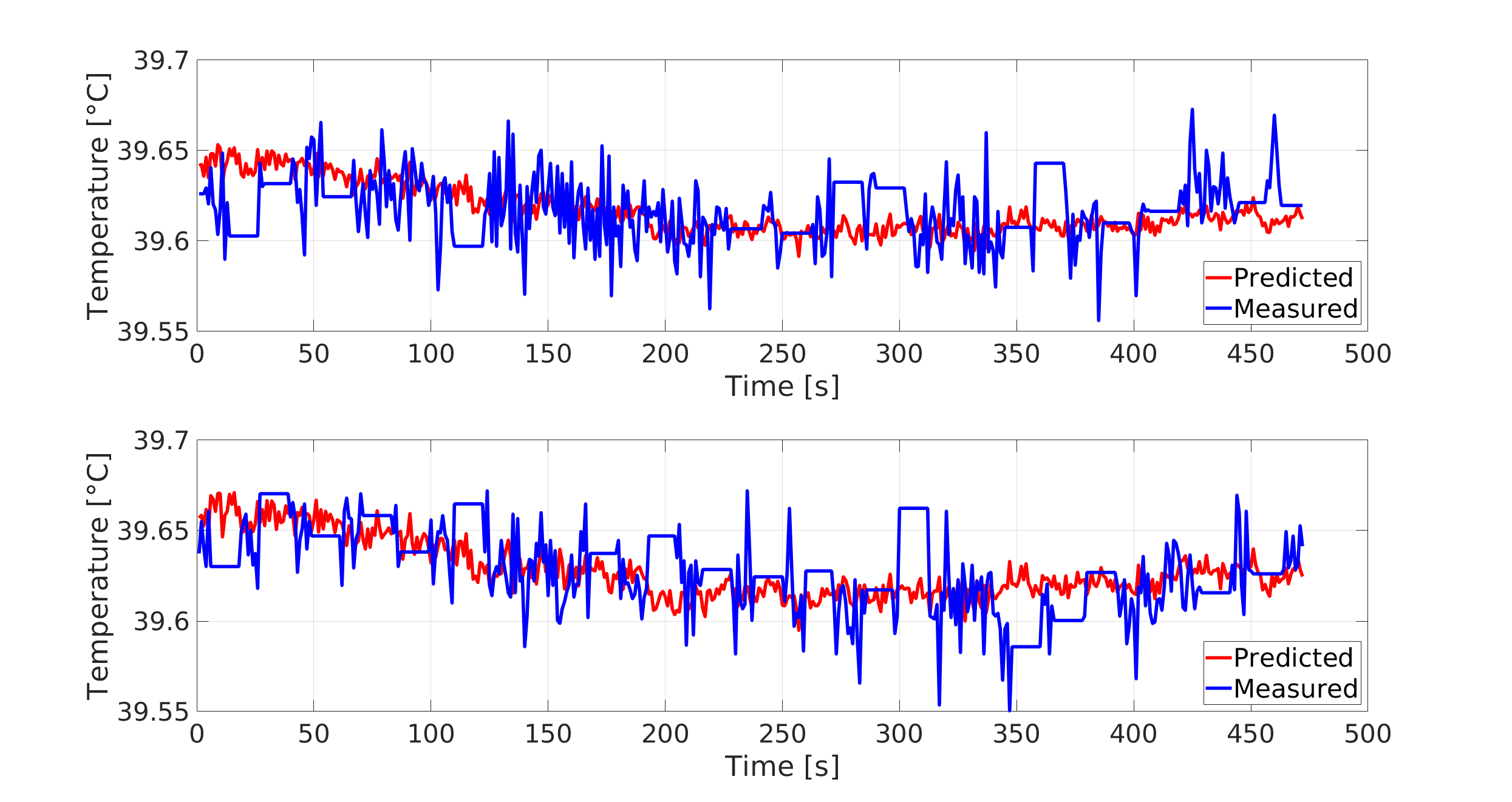}
		\caption{Motor 5: Different predictions with \textbf{unseen} joint torques.}
	\end{subfigure}%
	\begin{subfigure}[t]{0.5\textwidth}
		\centering
		\includegraphics[height=2.1in, width=\linewidth, keepaspectratio]{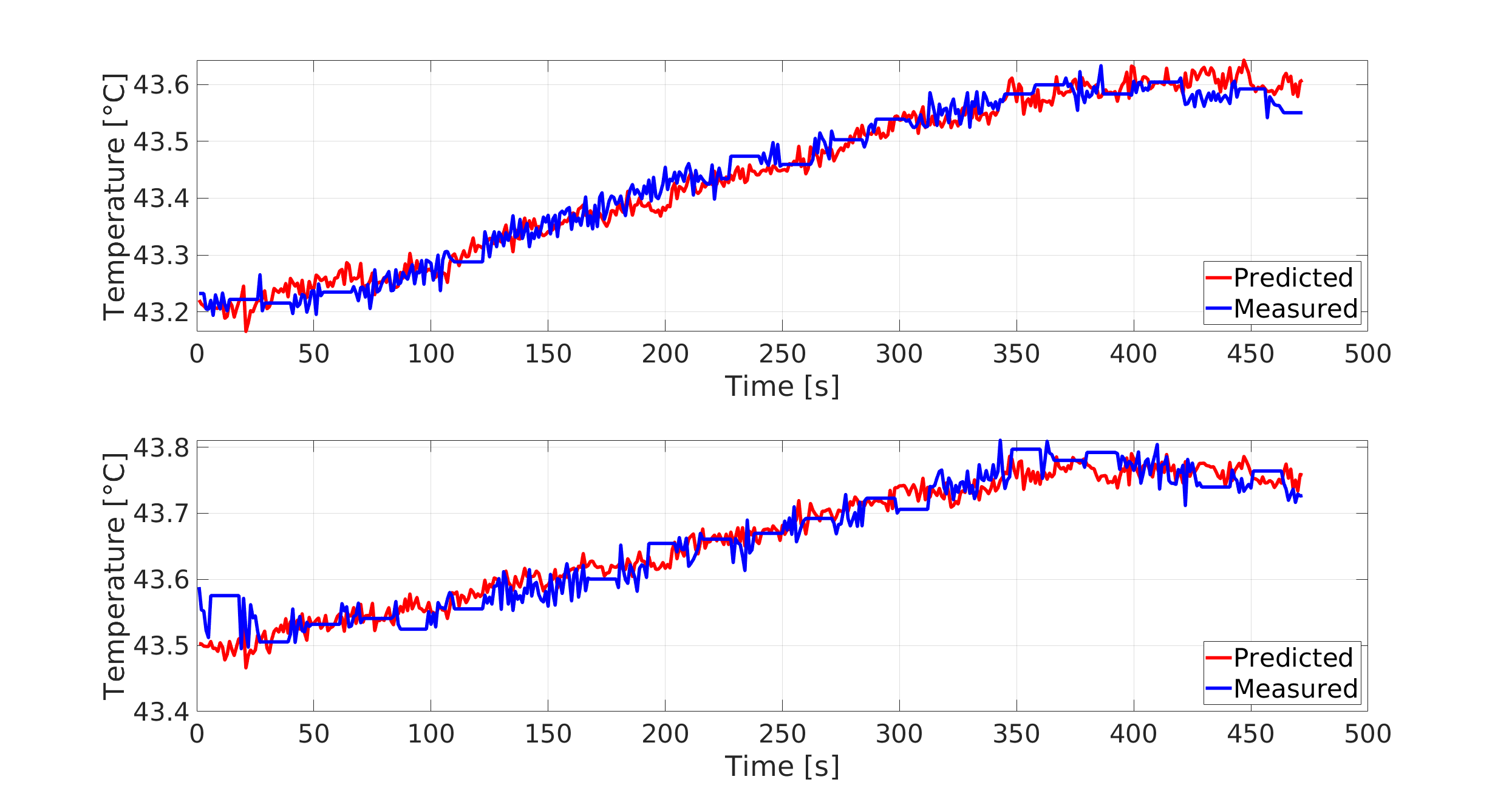}
		\caption{Motor 6: Different predictions with \textbf{unseen} joint torques.}
	\end{subfigure}

    \vspace{1ex} 

	\begin{subfigure}[t]{0.5\textwidth}
		\centering
		\includegraphics[height=2.1in, width=\linewidth, keepaspectratio]{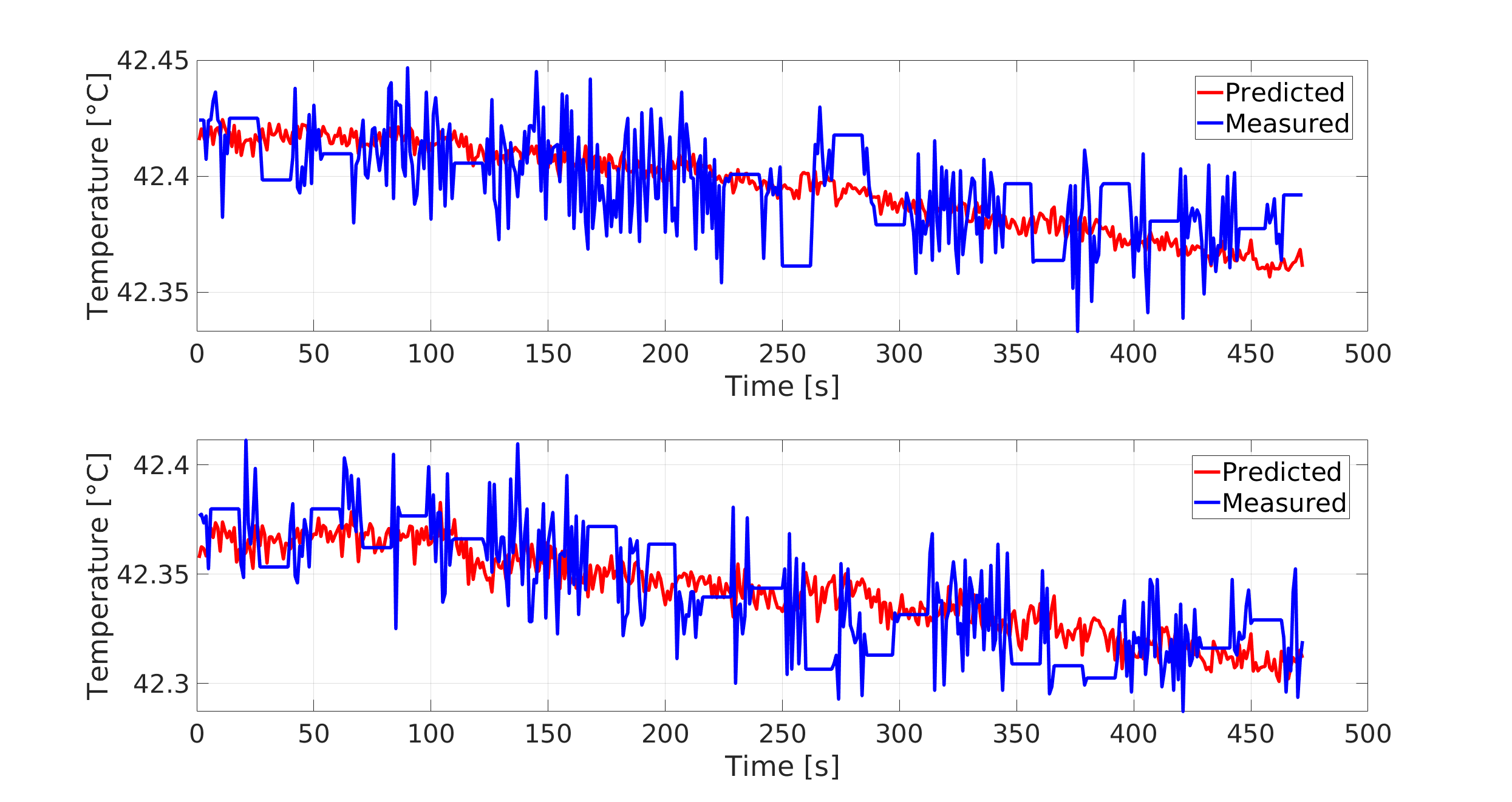}
		\caption{Motor 7: Different predictions with \textbf{unseen} joint torques.}
	\end{subfigure}%
	\begin{subfigure}[t]{0.5\textwidth}
		\centering
		\includegraphics[height=2.1in, width=\linewidth, keepaspectratio]{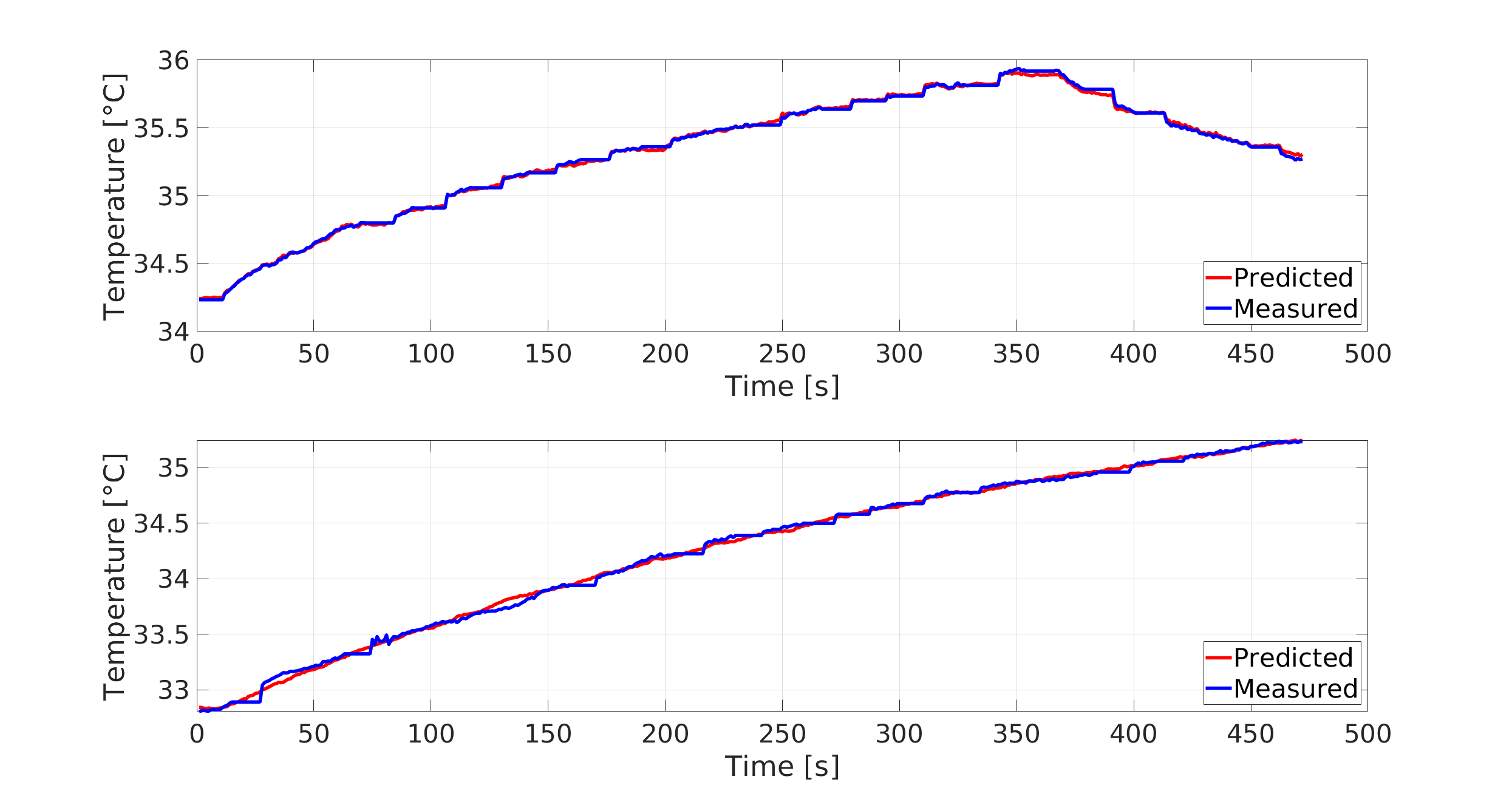}
		\caption{Motor 2: Different predictions with \textbf{seen} joint torques.}
	\end{subfigure}

	\caption{Capturing the thermal behavior of joint motors of the robot in Fig.~\ref{fig:intro} with previously unseen and seen data.}
	\label{sevenjoint} 
\end{figure*}

\section{Application}
Experiments have been carried out to predict the temperature of all the seven joint motors of the robot in Fig.~\ref{fig:intro} by using joint torques and without any  information  about how the joint torques map to motor temperatures and any assumption on the joint torque profiles. Another goal impacting the energy autonomy  was to keep training data  for a satisfactory prediction accuracy small. We target remote use cases \mbox{with on-board intelligence in the future.}

\subsection{Data collection and pre-processing}
The collected datasets allow the utilization of joint positions, torques, velocities, and currents of all seven joints as input features.
The selection of the input values for the training can be adjusted as needed, varying from 7 to a total of 28 inputs (Fig.~\ref{fig:DataAquisition}). However, the best result was observed with only the joint torques (see Fig.~\ref{jointtorque}) as input features for the neuronal network. The temperature values of the seven joints are used as targets.
To evaluate the temperature predictions from the neural networks, randomly selected temperature measurements serve as ground truth.

The setup for collecting joint torques and corresponding motor temperatures is shown in Fig. \ref{fig:DataAquisition}. Recorded profiles of joint torques are rather challenging when compared to step profiles considered in related works.  Acquired data are not filtered. A z-score normalization of  input (i.e., joint torques) and target (motor temperatures) data is carried out before training. The  mean $\mu$ and standard deviation $\sigma$ of target temperature data from the normalization are stored. After training, the original  temperature data $x$ is reconstructed from the predicted $\tilde{x}$ according
\begin{equation}
	x=\tilde{x}\sigma +\mu
\end{equation}
Data normalization has significantly sped up the training process and enhanced the prediction accuracy (see Fig.~\ref{trainingloss}). 
\subsection{Model training}
The mean squared error between network output and target temperature data  is adopted as loss function. Back propagation is leveraged for the gradient-based optimization of network weights. The Adaptive Moment {Estimation adapts the learning rate.} The probability with which input elements are dropped out (i.e., set to zero) to prevent  network over-fitting is set to 0.1. A mixture of \textit{tanh}, \textit{elu}, and \textit{sigmoid} are distributed as activation functions (a type per layer) to inject non-linearities  into the network. The output layer activation is  {the identity  function.}
\subsection{Prediction results}
The trained neural network of seven hidden layers (one LSTM layer, six feed forward  layers with a decreasing number of neurons toward the output) is fed with torque data. The network input is a  dataset \textit{unseen} during training.  The goal is to  assess the generalization capability of the network when facing new dynamics of the robot (Fig.~\ref{sevenjoint}). Two generalization tests are completed for each of the seven joints.  An additional test with seen data is conducted and restricted to the second joint. Validation and generalization results are shown in Fig.~\ref{sevenjoint}. Table~\ref{tb:rmseMaxAE} summarizes the RMSE and the max. absolute error (MaxAE) of the predictions. 
Whereas motor temperatures are predicted with negligible RMSE (below $0.17^\circ$) 
when the two seen  datasets are used as inputs of the trained network (see Fig.~\ref{sevenjoint}.h), the absolute value of the overall prediction error remains below $0.5^\circ$ in the case of totally unseen data (see  Fig.~\ref{sevenjoint}.a-g). 
It is worth noting that the small training dataset is randomly generated. Generalization accuracy can be further enhanced by augmenting the amount of training data. However, an advantage of working on a small dataset is to assess how a minimum amount of data, energy demand, and a satisfactory accuracy relate. In our specific case, 16 datasets with recording time length as in Fig.~\ref{jointtorque} is suitable. Insights gained are useful to steer on-board machine learning in  mobile \mbox{robots under energy autonomy constraints.}
\begin{table}[t!]
  \begin{center}
  \caption{RMSE and MaxAE of Fig.~\ref{sevenjoint} a-h.The first RMSE and MaxAE values refer to the upper, the second values refer to the lower plots of the respective motor temperatures.}\label{tb:rmseMaxAE}
  \resizebox{0.8\columnwidth}{!}{  
  \begin{tabular}{|l|c|c|}
  \hline
  {Motor no.} & {RMSE [$^\circ$C]} & {MaxAE [$^\circ$C]} \\
  \hline
  (a) Motor 1 & 0.0076 $|$ 0.0068 & 0.0246 $|$ 0.0224 \\
  (b) Motor 2 & 0.0853 $|$ 0.1166 & 0.2674 $|$ 0.3503 \\
  (c) Motor 3 & 0.0094 $|$ 0.0105 & 0.0308 $|$ 0.0274 \\
  (d) Motor 4 & 0.1054 $|$ 0.1625 & \textbf{0.3327} $|$ \textbf{0.4275} \\
  (e) Motor 5 & 0.0202 $|$ 0.0209 & 0.0638 $|$ 0.0826 \\
  (f) Motor 6 & 0.0259 $|$ 0.0275 & 0.0762 $|$ 0.0974 \\
  (g) Motor 7 & 0.0178 $|$ 0.0188 & 0.0441 $|$ 0.0535 \\
  (h) Motor 2 (seen) & 0.0152 $|$ 0.0290 & 0.0467 $|$ 0.0911 \\
  \hline
  \end{tabular}
  }
  \end{center}
\end{table}

\section{Conclusion}
This work shows that the temperature behavior of a robot can be predicted in advance without any analytical  model about how joint torques map to motor temperatures. The proposed approach is not only model-free but also scales up for any number of joint motors and generalizes when exposed to  unseen inputs despite the small amount of training data being used. Its effectiveness is evaluated and demonstrated for complex joint torque profiles and  each of the seven joints of the experimental robot. The proposed framework can help anticipate and alleviate widespread shutdown  of industrial and service robots once a critical temperature  is attained, design thermally uncritical joint trajectories, prolong the lifetime of robot motors, and prevent thermal burns in industry and society.


\bibliography{ifacconf}             
                                                   







\appendix
\end{document}